# Comparison of Artificial Intelligence Techniques for Project Conceptual Cost Prediction

**Haytham H. Elmousalami**
Project engineer and PMP at General Petroleum Company (GPC), Egypt.
MS.c, Department of Construction and Utilities, Faculty of Engineering, Zagazig University, Egypt.
E-mail: Haythamelmousalami2014@gmail.com

**Abstract:** Developing a reliable parametric cost model at the conceptual stage of the project is crucial for projects managers and decision makers. Existing methods, such as probabilistic and statistical algorithms have been developed for project cost prediction. However, these methods are unable to produce accurate results for conceptual cost prediction due to small and unstable data samples. Artificial intelligence (AI) and machine learning (ML) algorithms include numerous models and algorithms for supervised regression applications. Therefore, a comparison analysis for AI models is required to guide practitioners to the appropriate model. The study focuses on investigating twenty artificial intelligence (AI) techniques which are conducted for cost modeling such as fuzzy logic (FL) model, artificial neural networks (ANNs), multiple regression analysis (MRA), case-based reasoning (CBR), hybrid models, and ensemble methods such as scalable boosting trees (XGBoost). Field canals improvement projects (FCIPs) are used as an actual case study to analyze the performance of the applied ML models. Out of 20 AI techniques, the results showed that the most accurate and suitable method is XGBoost with 9.091% and 0.929 based on Mean Absolute Percentage Error (MAPE) and adjusted $R^2$. Nonlinear adaptability, handling missing values and outliers, model interpretation and uncertainty have been discussed for the twenty developed AI models.
**Keywords:** Artificial intelligence, Machine learning, ensemble methods, XGBoost, evolutionary fuzzy rules generation, Conceptual cost, and parametric cost model.

## Introduction

Conceptual cost estimate occurs at 0% to 2% of the project completion where limited information about the project is available with a high level of uncertainty and unknown risks (Hegazy and Ayed, 1998). Conceptual cost prediction is considered one of the fundamental criteria in the projects' decision making and feasibility studies at the early stages of the project. The estimating should be completed within a limited time period. Therefore, the accurate conceptual cost estimate is a challengeable task for cost engineers, project managers, and decision makers (Jrade 2000). Parametric cost modeling is to develop a model based on logical or statistical relations of the key cost drivers extracted by conducting qualitative techniques (ElMousalami et al. 2018 a) or statistical analyses such as factor analysis (Marzouk, and Elkadi 2016) or stepwise regression technique (ElMousalami et al. 2018 b). The main motivations to automate cost estimation are that:

1. Quantity survey is time and effort consuming process (Sabol, 2008).
2. Cost estimation may prone to human errors during estimation or personal judgment where biases and inaccuracy can exist.
3. High accurate and reliable tool is required for project managers and decision makers.

Artificial intelligence (AI) includes powerful techniques to automate cost estimate with high precision based on collected projects data. However, the accuracy of cost prediction is a major criterion in the success of any construction project, where cost overruns are a critical unknown risk, especially with the current emphasis on tight budgets. Moreover, cost overruns can lead to the cancellation of a project (Feng et al. 2010; AACE 2004). Therefore, improving the prediction accuracy is the main requirement in developing the cost model. Small data size, missing data values, maintaining uncertainty, computational complexity and model interpretation are the key challenges during prediction modeling. Accordingly, what are the best technique to model the project conceptual cost?

***Research Methodology***

The objective of the study is to answer the last question through conducting a comparative analysis to AI techniques for conceptual cost modeling. The scope of this study focuses on the most common (AI) techniques such as supportive vector machine (SVM), fuzzy logic (FL) model, artificial neural networks (ANNs), Multiple regression analysis (MRA), case-based reasoning (CBR), hybrid models, and ensemble methods such as scalable boosting trees (XGBoost), diction tree (DT), random forest (RF), Adaboost, scalable extreme gradient boosting machines (XGBoost), and evolutionary computing (EC) such as genetic algorithm (GA).

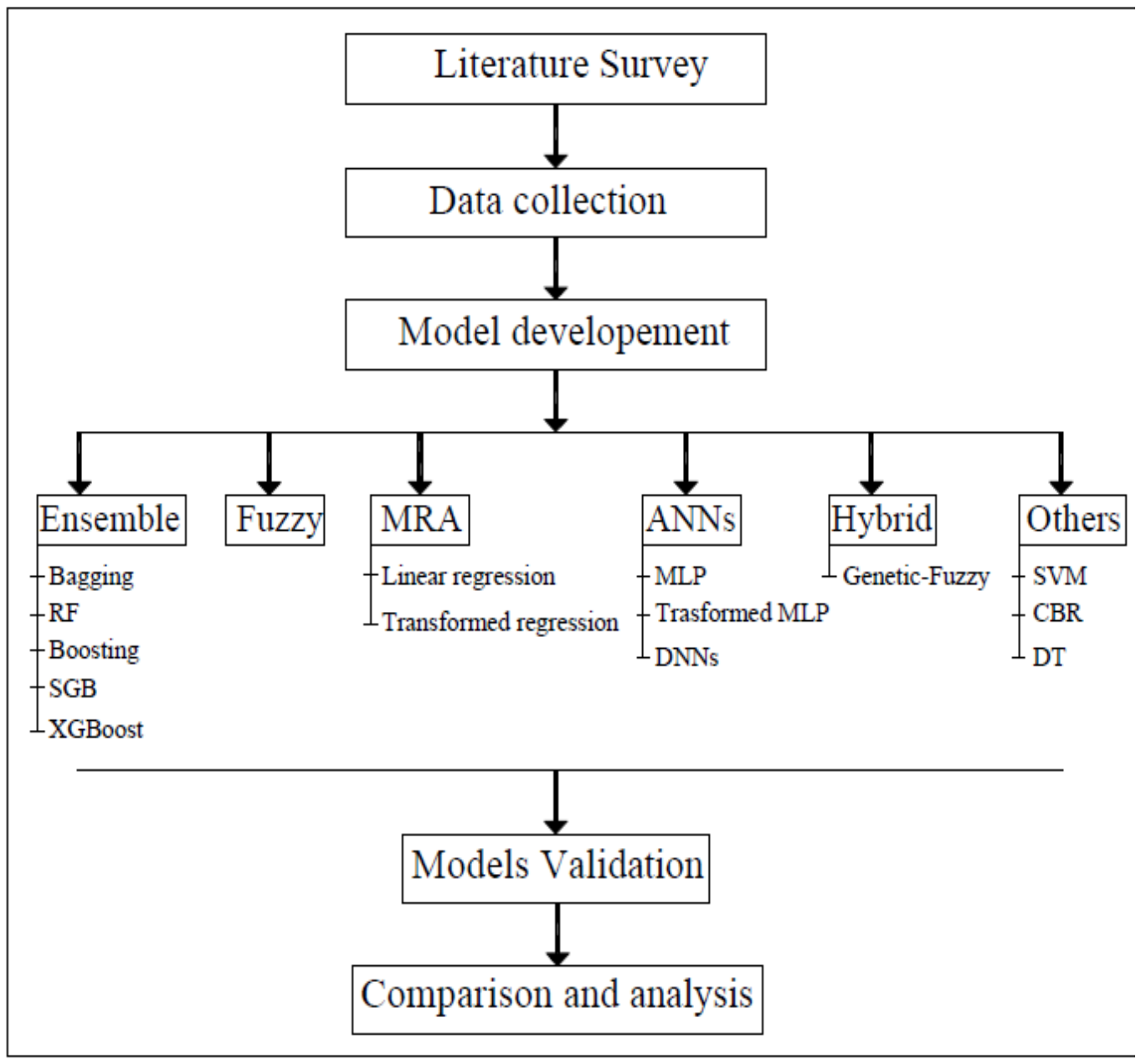

Fig.1.Research methodology.

As illustrated in Fig.1, the first step in the proposed methodology is a literature review of the previous practices. The second step is data collection of FCIPs historical cases. The third step is model development based on the AI techniques for cost prediction. The fourth step in model validation and the final step is models comparisons and analysis.

**Literature review**

Many previous studies have applied AI techniques and ML models. A semilog regression model has performed to develop cost models for residential building projects in German with a prediction accuracy of 7.55% (Stoy et al, 2012). Based on 92 building projects, ANNs and SVM have been used to predicted cost and schedule success at the conceptual stage. Such a model has a prediction accuracy of 92% and 80% for cost success and schedule success, respectively (Wang et al, 2012). Based on 657 building projects in Germany, a multistep ahead approach is conducted to increase the accuracy of the model's prediction (Dursun and Stoy, 2016). Marzouk and Elkadi (2016) have applied ANNs where the MAPE for test sets was 21.18%. Fan et al. (2006) have developed a decision tree approach for investigating the relationship between house prices and housing characteristics. Monte Carlo simulation and a multiple linear regression model have been developed as a benchmark model to evaluate the model's performance where the MAPE was 7.56. Wang and Ashuri (2016) have developed a highly accurate model based on random tree ensembles to predict construction cost index where the model's accuracy has reached 0.8%. Williams and Gong (2014) have built a stacking ensemble learning and text mining to estimate the cost overrun using the project contract document where the accuracy was 44%. Chou and Lin (2012) have established an ensemble learning model of ANNs, SVM and a decision tree for predicting the potential for disputes in public-private partnership (PPP) with the accuracy of 84%. Analytic hierarchy process (AHP) has incorporated into CBR to build a reliable cost estimation model for highway projects in South Korea (Kim, 2013). However, the main gap of these studies is developing deterministic predictive models without taking uncertainty nature into account where adding uncertainty nature to the predicted values improves the quality and reliability of the developed models (Zadeh, 1965, 1973).

Consequently, Fuzzy theory can be conducted to handle uncertainty concept to prediction modeling (Zadeh, 1965, 1973). Based on 568 Towers, a four input fuzzy clustering model and sensitivity analysis are conducted for estimating telecommunication towers with acceptable MAPE (Marzouk and Alarabyb, 2014). Shreenaath et al, (2015) have conducted a statistical fuzzy approach for prediction of construction cost overrun. The FL model is developed for satellite cost estimation. Such model works as a fuzzy expert tool for cost prediction based on two input parameters (Karatas and Ince, 2016). However, these studies have developed fuzzy systems without mentioning the method of fuzzy rules generation or the fuzzy rules has been developed based on experts' experience. Determining the fuzzy rules is the main gap of the previous studies. Therefore, a new trend evolves to solve this problem such as developing hybrid fuzzy modeling for the cost estimate purposes such as evolutionary-fuzzy modeling.

Zhai et al, (2012) have created an improved fuzzy system which is established based on fuzzy c-means (FCM) to solve the problem of fuzzy rules generation. Zhu et al, (2010) have conducted an evolutionary fuzzy neural network model for cost estimation based on eighteen examples and two examples of training and testing, respectively. Cheng and Roy, (2010) have developed a hybrid artificial intelligence (AI) system based on supportive vector machine (SVM), FL and GA for decision making construction management.

**Application to Field canals improvement projects (FCIPs)**

In this section, the selected AI techniques are applied to the conceptual cost prediction of FCIPs in Egypt as an actual case study.

***Case Background***

FCIPs are one of the main projects in Irrigation Improvement Projects (IIPs) in Egypt. The strategic aim of these projects is to save fresh water, facilitate water usage and distribution among stakeholders and farmers. To finance this project, conceptual cost models are important to accurately predict preliminary costs at early stages of the project (Elmousalami et al. 2018 b; Radwan, 2013).

***Data collection and feature selection***

Elmousalami et al. (2018 a) have conducted qualitative approachs such as Fuzzy Delphi method and fuzzy analytical hierarchy process to rank the cost drivers. Moreover, Elmousalami et al. (2018 b) have developed a quantitative hybrid approach based on both Pearson correlation and stepwise regression to filter the key cost drivers. The key cost drivers were area served (P1), pipeline total length (P2), and the number of irrigation valves (P3), and construction year (P4). Accordingly, a total of 144 FCIPs during 2010 and 2015 have been Collected. For validation purposes, this collected sample has randomly branched into a training sample (111 instances) and a testing sample (33 instances).The training sample in the present case study is 111 instances would be sufficiently acceptable to train reliable ML models where Green (1991) concluded that [50 + 8*N] is the minimum sample size, and N is the number of independent variables (key cost drivers).

**Artificial intelligence (AI) techniques developments**

AI techniques are aspects of human knowledge and computational adaptively to become more vital in system modeling than classical mathematical modeling (Bezdek 1994). Based on AI, an intelligent system can be developed to produce consequent outputs and actions depending on the observed inputs and outputs of the system (Siddique and Adeli 2013; Bishop 2006).

***Case-based reasoning (CBR)***

Case-based reasoning (CBR) is a sustained learning and incremental approach that solves problems by searching the most similar past case and reusing it for the new problem situation (Aamodt and Plaza 1994). Therefore, CBR mimics a human problem solving (Ross 1989; Kolodner 1992). CBR is a cyclic process of learning from past cases to solve a new case. The main processes of CBR are retrieving, reusing, revising and retaining. The retrieving process is solving a new case by retrieving the past cases. The case can be defined by key attributes. Such attributes are used to retrieve the most similar case, whereas, reusing process is utilizing the new case information to solve the problem. Revising process is evaluating the suggested solution to the problem. Finally, retaining process is to update the stored past cases with such a new case by incorporating the new case to the existing case-base (Aamodt and Plaza 1994). A CBR model is developed to predict the conceptual cost of FCIP based on similarity attribute of the entered case comparable with the stored cases. Once attributes are entered, attributes similarities (AS) can be computed based on equation (1) (Kim & Kang, 2004).

$$\text{AS} = \frac{\text{Min}(AV_N, AV_R)}{\text{Max}(AV_N, AV_R)} \qquad (1)$$

Where AS = Attribute Similarity, $AV_N$ = Attribute value of new entered case, $AV_R$ = Attribute value of retrieved case. Depending on AS and attribute weights (AW), case similarity (CS) can be computed by equation (2) (Perera and Waston, 1998). AW are selected by an expert to emphasize the existence and importance of the case attributes.

$$\text{CS} = \frac{\sum_{i=1}^{n}(AS_i \, * \, AW_i)}{\sum_{i=1}^{n}(AW_i)} \qquad (2)$$

Where CS is case similarity, AS is attribute similarity, AW = attribute weight, (i) is the number of attributes (key cost drivers).

***Fuzzy logic model***

Fuzzy logic (FL) is to model human reasoning taking uncertainties possibilities into account where incompleteness, randomness, and ignorance of data are represented in the model (Zadeh, 1965, 1973). If–Then rule statements are utilized to formulate the conditional statements that develop FL rules base system. As shown example in Fig(2), there are two parameters $X_1$ and $X_2$ where $\mu X_1 = \{a_1, b_1, c_1, d_1\}$, $\mu X_2 = \{a_2, b_2, c_2, d_2\}$, $\mu Y = \{a_y, b_y, c_y, d_y\}$ and the fuzzy system consists of two rules as following:

*Rule 1: IF $x_1$ is $a_1$ AND $x_2$ is $c_2$ THEN y is $a_y$.*

*Rule 2: IF $x_1$ is $b_1$ AND $x_2$ is $d_2$ THEN y is $b_y$.*

Where two inputs are used $\{X_1=4, X_2=6\}$. Such two inputs intersect with the antecedents MF of the two rules where two consequents rules are produced $\{R_1$ and $R_2\}$ based on minimum intersections. The consequent rules are aggregated based on maximum intersections where the final crisp value is 3. The aggregated output for $R_i$ rules are given by

*Rule 1: $\mu R_1 = \min [\mu a1 (x1) \text{ and } \mu c2 (x_2)]$*

*Rule 2: $\mu R_2 = \min [\mu b1 (x1) \text{ and } \mu d2 (x_2)]$*

*Y: Fuzzification [max [$R_1$, $R_2$]*

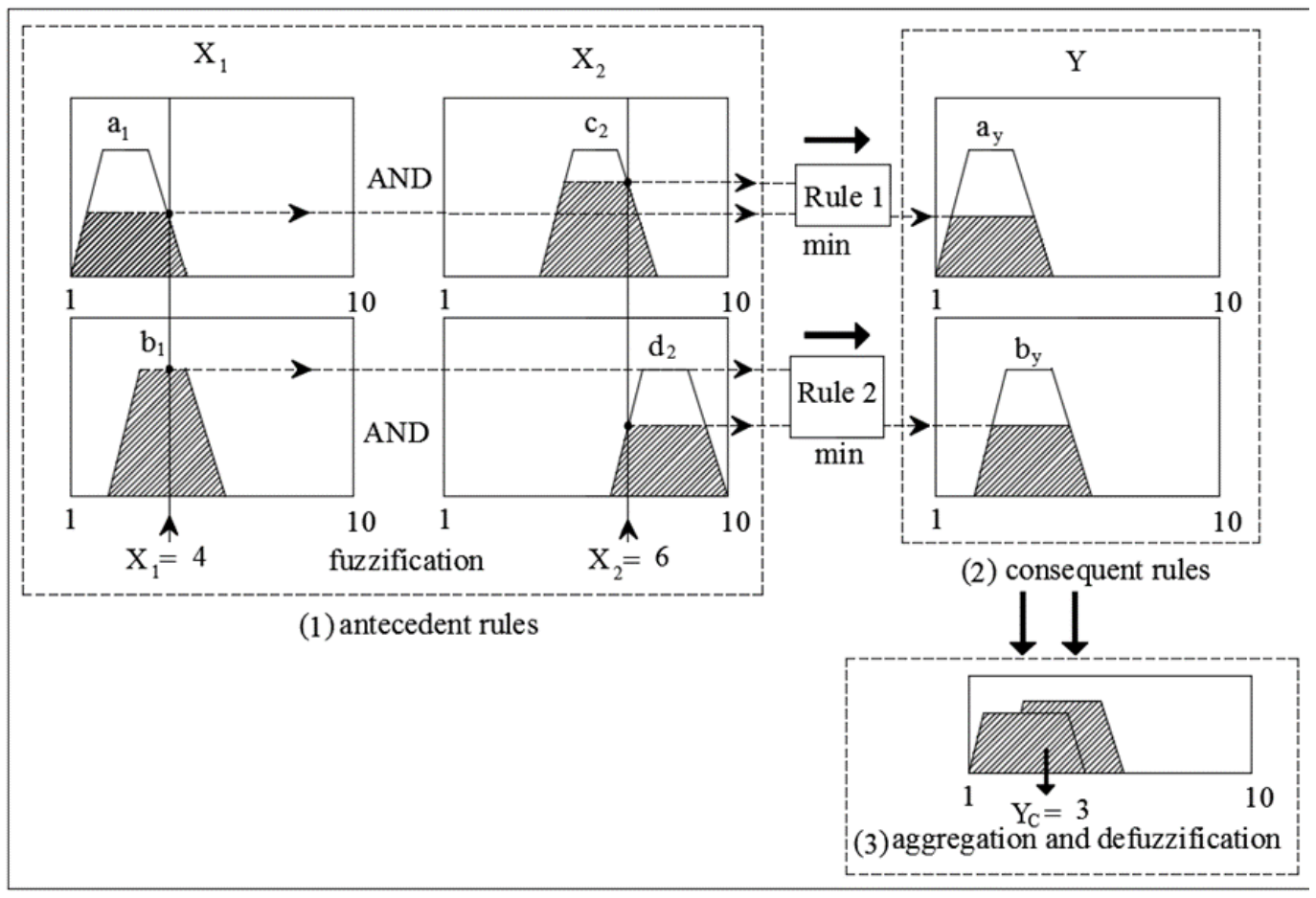

Fig. 2. Fuzzy rules firing.

The first step in the FL model is fuzzification the four key cost drivers and identify their MFs. The most critical stage is to develop fuzzy rules base. experts are consulted to give their experience to develop such rules. As shown in Fig. 3, seven triangle MFs have been used to fuzzify the variables of FCIPs For example, the input variable construction year consists of seven triangle MFs {$MF_1$, $MF_2$, $MF_3$, $MF_4$, $MF_5$, $MF_6$, $MF_7$}. Accordingly, the number of possible rules equals $7^4$ rules. Therefore, there is a need to automatically generate such rules. For FL model, a total of 190 IF-Then rules have been formulated.

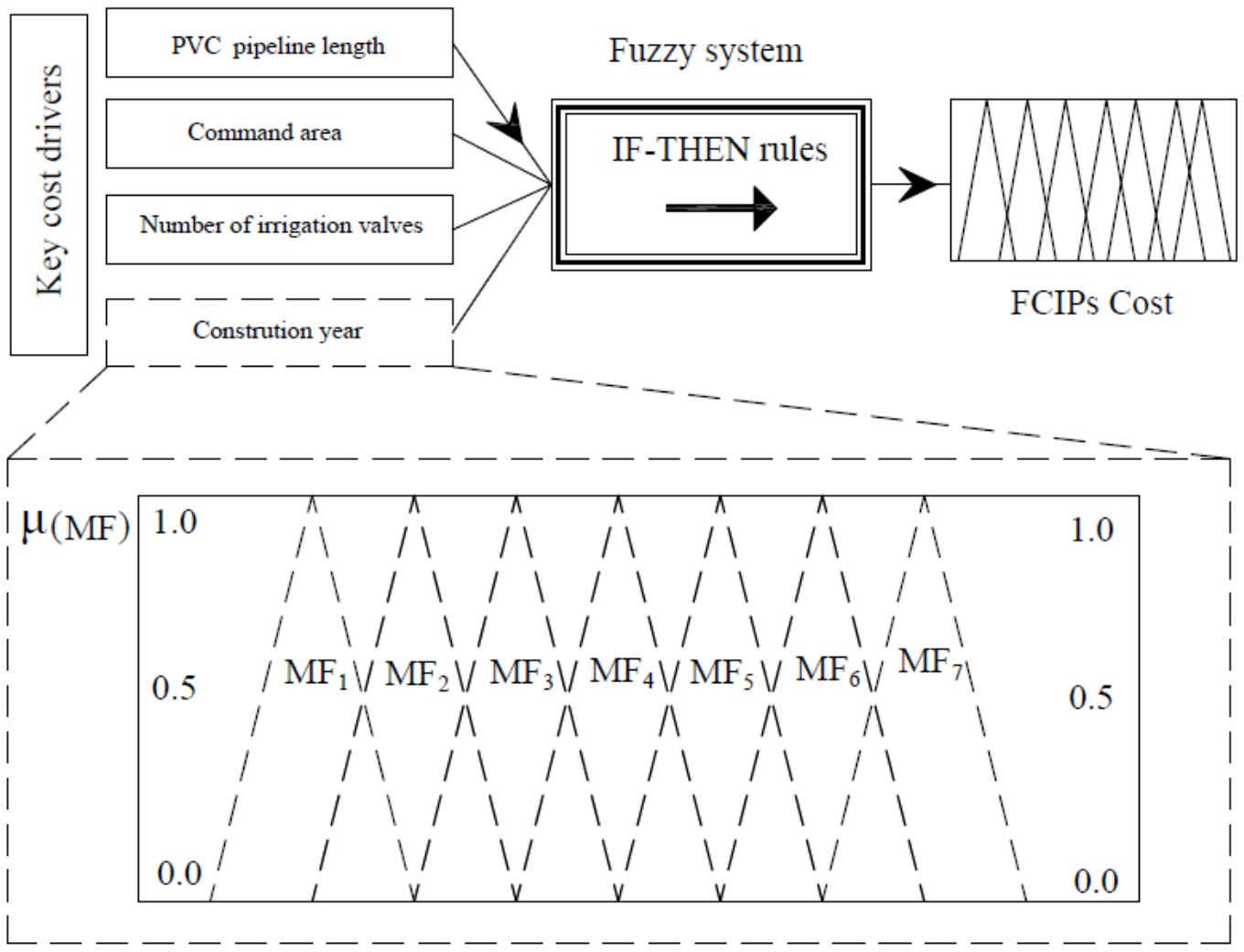

***Fig. 3. Fuzzy logic model for FCIPs.***

***Genetic-Fuzzy model***

Many approaches exist for evolutionary fuzzy hybridization [(Angelov, 2002); (Pedrycz et al, 1997)]. Traditionally, an expert is consulted to define such fuzzy rules or the fuzzy designer can use trial and error approach to map the fuzzy rules and MFs. However, such an approach is time-consuming and does not guarantee the optimal set of the fuzzy rules. Moreover, the number of fuzzy IF-Then rules increase exponentially by increasing the number of inputs, linguistic variables, or a number of outputs. In addition, the experts cannot easily define all required fuzzy rules and the associated MFs. In many engineering problems, the evolutionary algorithm (EA) has been conducted to automatically develop fuzzy rules and MFs to improve the system performance (Chou 2006; Loop et al. 2010).

Genetic-Fuzzy model has been developed to optimally generate fuzzy rules. The study has applied Genetic algorithm (GA) to optimally select the fuzzy rules where 2401 rules represent the whole possible search space for GA. The formulation of genetic algorithm model depends mainly on defining two core terms: a chromosome representation and an objective function. First, based on Michigan approach, the chromosomes represents the fuzzy rules where the number of chromosomes ($CH_n$) are the number of fuzzy rules. Each chromosome is consists of five genes where four genes are for the key cost drivers the fifth gene is for the output (the cost of FCIP). Each gene consists of one of the seven membership functions ($MF_i$) where ($i$) is ranging from one to seven *($MF_1$:$MF_7$)* as shown in Fig.4. For example: **IF** {Area served ($P_1$) is $MF_5$ **AND** Total length ($P_2$) is $MF_2$ **AND** Irrigation valves ($P_3$) is $MF_2$ **AND** construction year ($P_4$) is $MF_6$} **THEN** {The Cost LE / Mesqa is $MF_3$}.

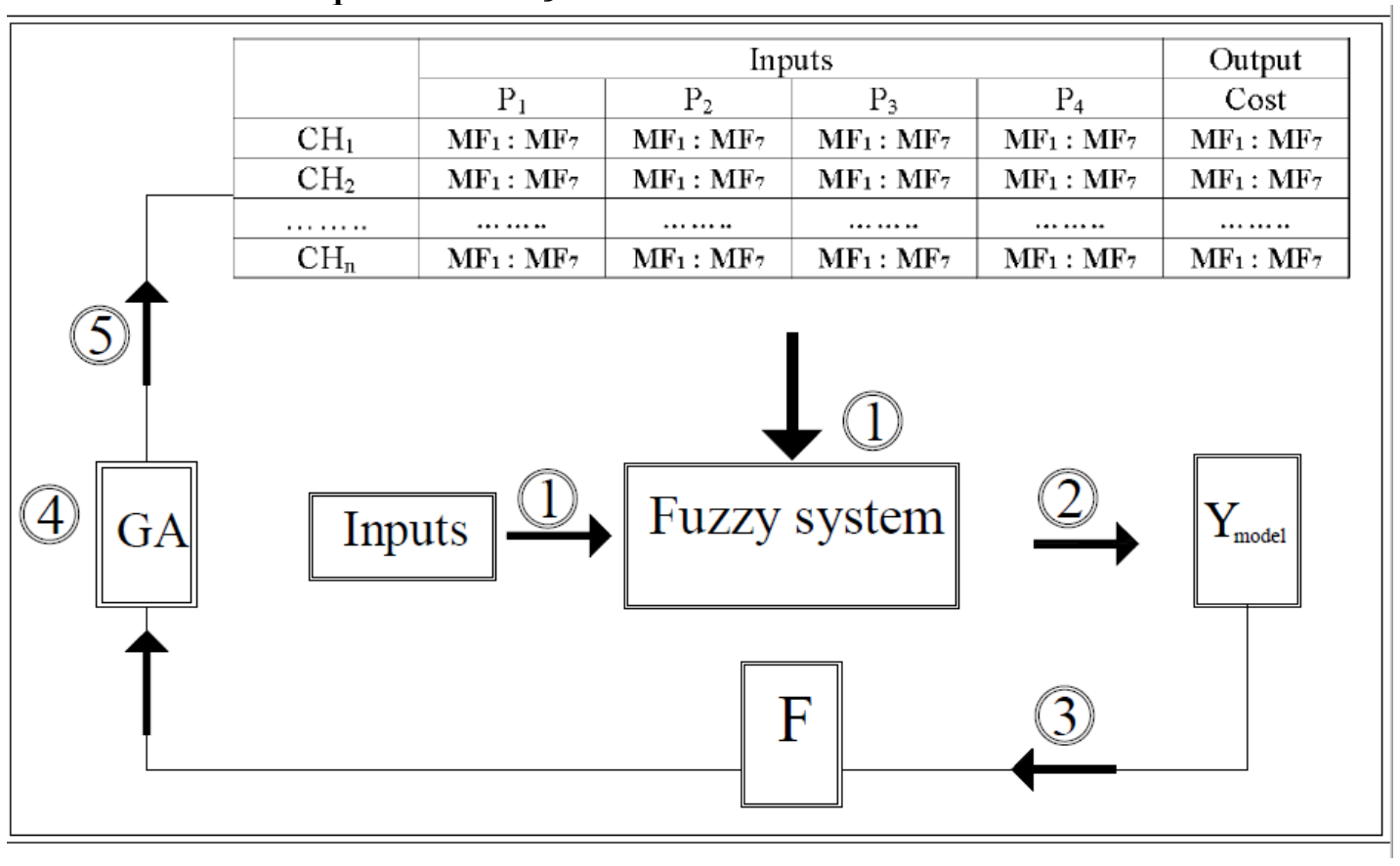

Fig. 4. The process of genetic fuzzy system.

Secondly, the fitness function is problem-dependent where the objective is to enhance the accuracy and quality of the system performance (Hatanaka et al., 2004). The fitness function is formulated to minimize the MAPE as equation (3).

$$\boldsymbol{F = MAPE = \frac{1}{n}\sum_{i=1}^{n}\frac{|y_i - \hat{y}_i|}{\hat{y}_i} x100} \qquad (3)$$

Where: (F) is a fitness function, (n) is the number of cases, (i) is the number of the case and $\hat{y}_i$ is the outcome of the model and $y_i$ is the actual outcome. As shown in Fig.4, the process of the developed model consists of five main steps:

1. An initial population of chromosomes has been identified to represent the initial state of the fuzzy rules. The four key cost drivers have been fed to the fuzzy system.
2. The fuzzy system produces the final output of the system $\hat{y}_i$
3. The $\hat{y}_i$ has been fed to fitness function (F) to evaluate the model performance.
4. GA uses the fitness function (F) to evaluate the search process where crossover probability and mutation probability have been set at 0.7 and 0.01 respectively.
5. The new population of fuzzy rules has been produced based on crossover and mutation processes.

***Support Vector Machines (SVM)***

SVM is a non-parametric supervised ML algorithm that can be applied for regression and classification problems (Vapnik 1979).

The study has applied the radial base function (RBF) as a kernel for Supportive vector regression model. A positive slack variable ($\xi$) will be added to handle the non-linearity of the data as the following equation (4) (Cortes and Vapnik 1995).

$$y_i\,(W.X_i + b) \geq 0 - \xi \quad ,\text{i} = 1,2,3, \ldots\ldots \text{ m} \qquad (4)$$

The objective is to minimize misclassifications cases through optimizing the margin and hyperplanes distance as equation (5).

$$Min \sum_{i=0}^{i=m} \frac{1}{2}\, w.w^T + C \sum_{i=0}^{i=m} \xi_i \qquad (5)$$

For i = 1,2,3, …… m where m is the number of cases.

***Decision trees (DT)***

DT is a supervised ML model that divides the cost data into hierarchical rules on each tree node by a repetitive splitting algorithm (Berry and Linoff 1997; Breiman et al. 1984). Classification and regression trees (CART) is a DT model that can be applied for both regression (continuous variables) and classification (categorical variables) applications (Quinlan 1986). CART has been developed to the FCIPs data. The features of the tree are the key cost drivers and the terminal tree nodes (leaf nodes) are continues values of the project cost.

***MRA and transformed regressions***

Haytham et al. (2018 b) have developed five regression models: standard linear regression, quadratic model, reciprocal model, semilog model, and power model. The most accurate model is the quadratic model where the quadratic model is a dependent variable transformation by taking the Square Root (Sqrt). The regression model consists of four key cost drivers as independent variables and (Y) represent FCIP cost per field canal as the dependent predictor. The quadratic regression model is formulated as the following Equation (6):

$$\mathbf{(Y)^{0.5} = -37032.81 + 2.21{*}P_1 + 0.1691{*}P_2 + 2.265{*}P_3 + 18.594{*}P_4} \qquad \mathbf{(6)}$$

***ANNs and DNNs***

ANNs is a computational method that is inspired by neuron cells. The major advantage of ANNs is their ability to fit nonlinear data (Siddique and Adeli 2013). Haytham et al. (2018 b) have developed three ANNs models with structure (4-5-0-1). four represents the number of inputs (four key cost drivers), five represents the number of hidden nodes in the first hidden layer, zero means no second hidden layer used and one represents one node to produce the total cost of the FCIPs. The first model is the untransformed model whereas the second model is transformed by the square root of the project cost. The third model is transformed by the natural log of the project cost. The type of training is batch, the learning algorithm is the scaled conjugate gradient and the activation function is hyperbolic tangent. A standard rectified linear unit (ReLU) is an activation function that can enhance the computing performance of ANNs (LeCun et al. 2015; Nair & Hinton, 2010). Mathematically, ReLU is defined as:

$$A = \begin{cases} X_i & , \quad if\ X_i \geq 0 \\ 0 & , \quad X_i < 0 \end{cases}$$

Deep neural networks (DNNs) has been developed to be investigated. The structure of DNNs model consists of three hidden layers where each hidden layer contains 100 neurons. The activation function is ReLU function. Accordinly DNNs's structure is (4-100-100-100-1).

***Ensemble methods***

Ensemble methods (fusion learning) are elegant data mining techniques to combine multiple learning algorithms to enhance the overall performance (Hansen and Salamon 1990). Ensemble methods can apply ML algorithms such as ANN, DT, and SVM which are called "base model or base learner" as inputs for ensemble methods. The concept behind ensemble methods can be illustrated as Fig.5 and mathematically as equation (7) (Chen and Guestrin 2016).

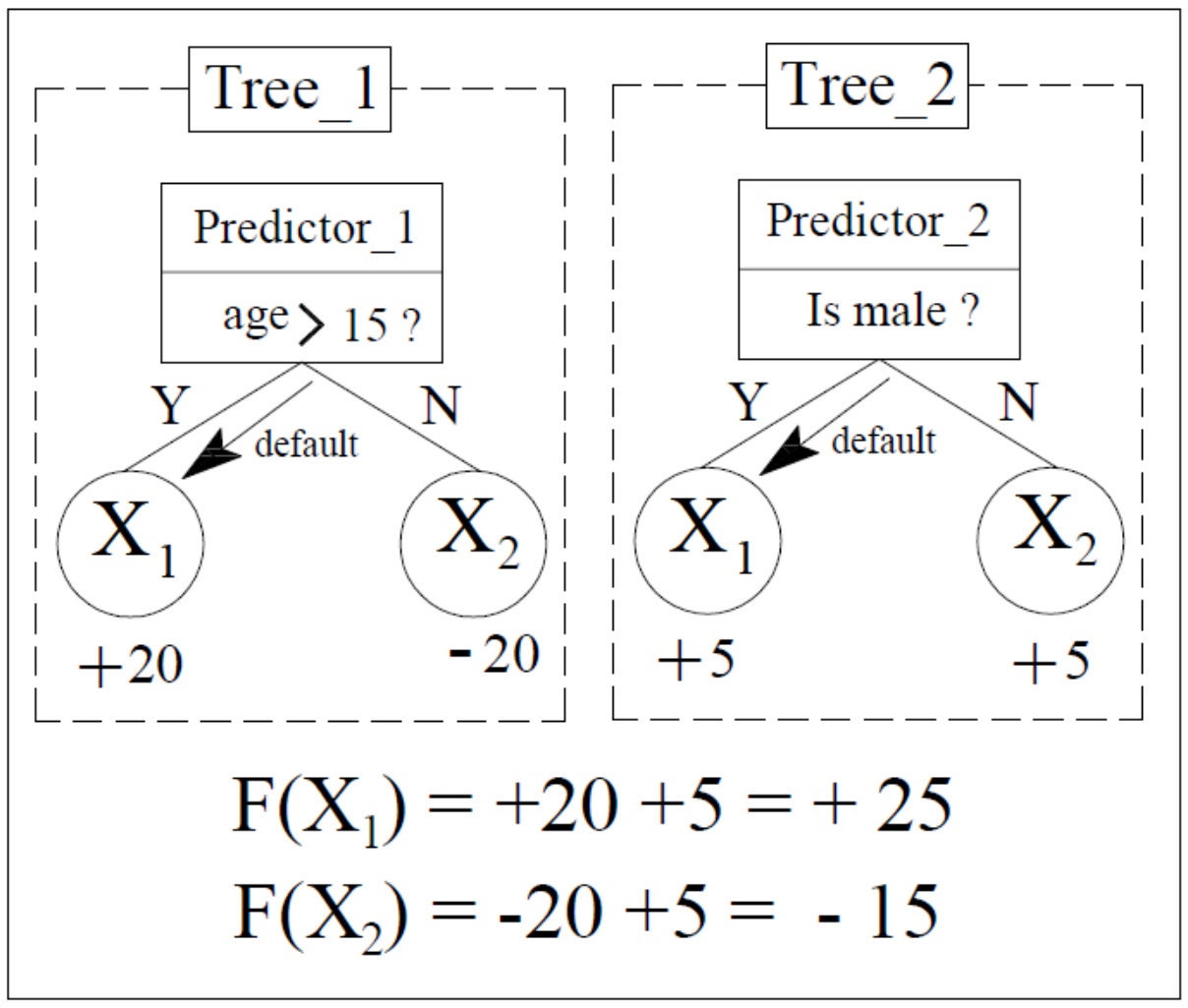

Fig. 5. Additive function concept.

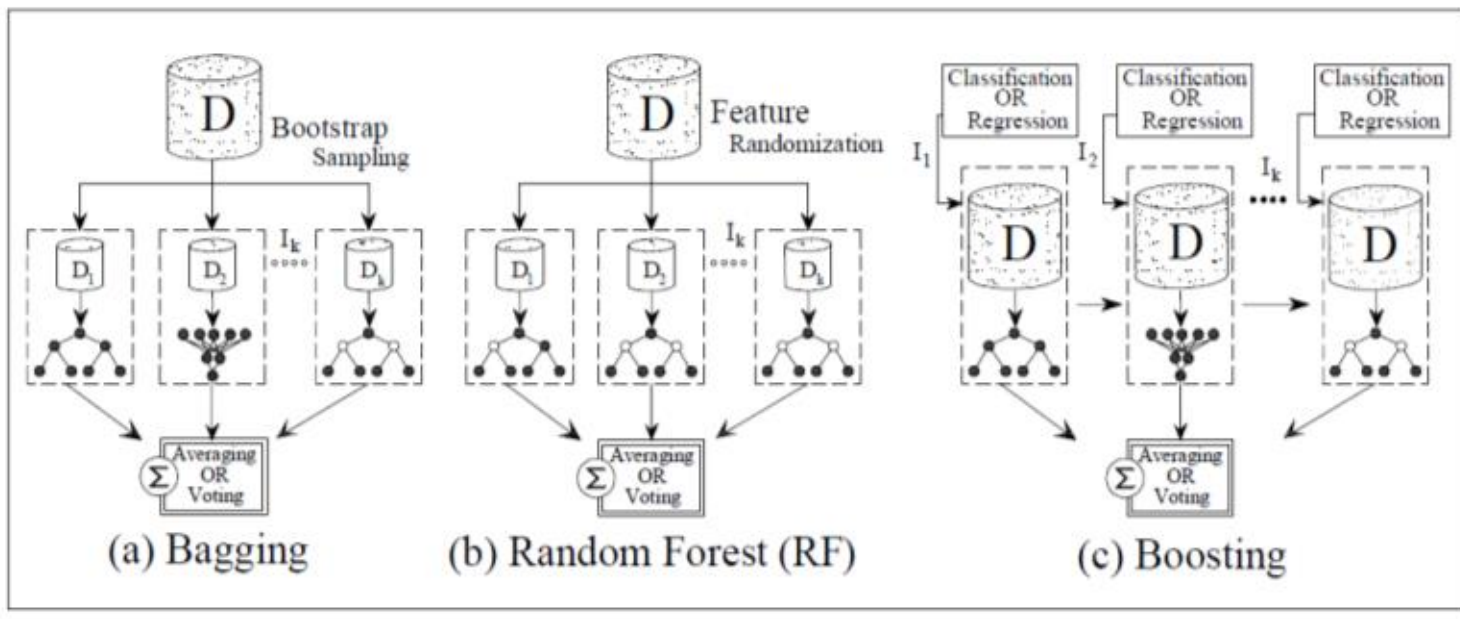

Fig. 6. (a) Bagging, (b) RF, and (c) Boosting.

For the given dataset (D) with *n* examples (144 cases) and *m* features (4 key cost drivers) D = {($x_i$; $y_i$)} ($x_i \in R^m$; $y_i \in R$) where R is the real numbers set. K is an additive function to predict the output as equation (7).

$$\hat{y}_i = \sum_{k=1}^{K} f_k\,(X_i) \quad , \quad f_k \in F \qquad (7)$$

Where F = { f (x) = $w_{q(x)}$ } (q: $\mathbb{R}^m \rightarrow$ T , T $\in$ $\mathbb{R}^m$). q is the structure of each tree that maps an example to the corresponding. T corresponds to the number of leaves in the tree. $\hat{y}_i$ is the predicted dependent variable (FCIPs cost LE / project). Each $f_k$ represents an independent tree structure q and leaf weights (w). ($X_i$) represents independent variables. (F) represents the regression trees space.

***Bagging***

Bagging is a variance reduction algorithm to train several classifiers based on bootstrap aggregating as shown in Fig.6 (a). Bagging algorithm randomly draws replicas of training dataset with replacement to train each classifier (Breiman 1996; Breiman 1999). As a result, diversity is obtained by resampling several data subsets. On average, each bootstrap sample contains 63.2% of the original training data set. The CART is selected as a base learner for bagging model.

***Random Forest (RF)***

Random Forests is a one of bagging ensemble learning models that can produce accurate performance without overfitting issue (Breiman 2001) as shown in Fig.6 (b). RF algorithms draw bootstrap samples to develop a forest of trees based on random subsets of features. Extremely randomized tree algorithm (Extra Trees) merges the randomization of random subspace to a random selection of the cut-point during splitting tree node process. Extremely randomized tree mainly controls the attribute randomization and smoothing parameters (Geurts 2006).

***Boosting and Adaptive boosting (Adaboost)***

Schapire has presented Boosting procedure (also known as adaptive resampling) as an algorithm that boosts the performance of weak learning algorithms (Schapire 1990). Bagging generates classifiers in parallel while boosting develops the classifiers sequentially as shown in Fig.6 (C). Thus, Boosting converts weak models to strong ones. Freund and Schapire (1997) have presented Adaptive Boosting algorithm (AdaBoost). AdaBoost is selected as one of the top ten data mining. AdaBoost serially manipulates the cost data for each base learner to AdaBoost assigns equal weights for all cases where larger weights are assigned to the misclassified cases. The objective is to make a greater focus on the misclassified cases to be corrected in the consequent iteration. In addition, AdaBoost algorithm assigns other weights to rank each individual base learning algorithm based on its accuracy (Bauer and Kohavi 1999). The CART is selected as a base learner for AdaBoost model to predict the conceptual cost of FCIPs.

***Extreme Gradient Boosting (XGBoost)***

XGBoost is a Large-scale ML system that can build a highly scalable end-to-end ensemble tree boosting system for big data processing (Chen and Guestrin 2016). XGBoost is a modified gradient tree model with regularization term to the additive function as equation (8):

$$\boldsymbol{L(\phi) = (x+a)^n = \sum_{k=0}^{n} l(\hat{y}_i, y_i) + \sum_{k=1}^{K} \Omega\,(f_k)}\,,$$

$$\textbf{where}$$

$$\boldsymbol{\Omega(f) = \gamma T + \frac{1}{2}\lambda\,\|w\|^2} \qquad (\mathbf{8})$$

*Where* $\boldsymbol{L}$ represents a differentiable convex cost function that determines the difference between the predicted output $\hat{\boldsymbol{y}}_{\boldsymbol{i}}$ and the actual output $y_i$. $\Omega$ is a regularization term to avoid overfitting and smooth the learned weights ($W_i$). The regularization term penalizes the complexity of the regression tree functions.

***Stochastic gradient boosting (SGB)***

The performance of gradient boosting can iteratively be improved as stochastic gradient boosting algorithm by injecting randomization into the selected data subsets. Injecting randomization to boosting algorithm can substantially boost both the fitting accuracy and computational cost of the gradient boosting algorithm [Breiman (1996); (Freund and Schapire, 1996)]. The training data is randomly drawn at each iteration without replacement from the data set. Stochastic gradient boosting can be viewed in this sense as a boosting bagging hybrid.

**Evaluation techniques**

Evaluation techniques for predictive models can be MAPE, the mean squared error (MSE), the root mean squared error (RMSE), the coefficient of determination ($R^2$) or adjusted $R^{*2}$

MAPE is comparing the predicted and actual outcomes (Makridakis et al. 1998) as equation (3). MAPE can be classified as an excellent prediction if MAPE is less than 10 %, between 10% from 20 % is good prediction. Between 20% to 50 % is acceptable forecasting and more than 50 % is inaccurate prediction (Lewis 1982). However, Peurifoy and Oberlender (2002) have defined 20% as acceptable limit for the conceptual cost estimate based on MAPE. Therefore, this study has categorized model accuracy to three main categories:

$$MAPE\ \%\ categorization = \begin{cases} below\ 10\ , & 10\% \geq MAPE \geq 0 \\ below\ 20\ , & 20\% \geq MAPE > 10\% \\ unacceptable\ , & MAPE > 20\% \end{cases}$$

Where “below 10” indicates high accuracy level than “*below* 20”.

The R-squared $R^2$ (coefficient of determination) is expressed as equation (9):

$$\boldsymbol{R^2 = 1 - \frac{SSE}{SST} = 1 - \frac{\sum_{i=1}^{n} (y_i - \hat{y}_i)^2}{\sum_{i=1}^{n} (y_i - \bar{y}_i)^2}} \qquad (\mathbf{9})$$

Where SSE is the sum of squares of the residuals and SST is the total sum of squares. $\bar{\boldsymbol{y}}_{\boldsymbol{i}}$ is the arithmetic mean of the Y variable. $R^2$ measures the percentage of the variation percentage of the predictor $\boldsymbol{y_i}$ explained by the dependent variable X. Thus, $R^2$ indicates how well the model fits the cost data. IF $R^2$ value of 0.9 or above, it is classified as very good, above 0.8 is good, above 0.5 is satisfactory, below 0.5 is poor (Aczel 1989; Ostertagová 2011). Adjusted R-squared $R^{*2}$ is computed by equation (10).

$$\boldsymbol{R^{*2} = R^2 - \frac{(1-R^2)K}{n-(K+1)}} \qquad (\mathbf{10})$$

Where $R^{*2}$ is adjusted for the number of variables included in the proposed model where $R^{*2}$ is lower than $R^2$ value. For model evaluation, $R^{*2}$ is always preferred to $R^2$ to avoid the over-fitting problem (Aczel 1989; Ostertagová 2011).

**Comparison and analysis**

MAPE and $R^{*2}$ have been validated the twenty developed models as displayed in Table.1. The whole developed models have been descendingly sorted from M 1 to M 20 based on MAPE as shown in Fig.7.

Elmousalami et at. (2018 b) have presented quadratic regression model (M 2) as the most accurate for FCIPs among the developed regression and ANNs models (M3, M4, M5, M6, M8, M13, and M14) with 9.120 and 0.851 for MAPE and R*2, respectively. However, this study presents that XGBoost (M 1) is more accurate than quadratic regression (M 2). XGBoost (M 1) comes in the first place slightly higher than M 2 with 9.091 % and 0.929 for MAPE and $R^{*2}$, respectively. Moreover, the unique advantage of the XGBoost is its high scalability where it can process noisy data and fit high dimension data without overfitting. XGBoost applies parallel computing to effectively reduce computational complexity and learn faster (Chen and Guestrin 2016). Another key advantage of XGBoost is handling the missing values where defaults direction is identified as shown in Fig.5. Accordingly, no effort is needed for cleaning the collected data (Fan, 2008).

Ensemble methods such as [Extra Trees (M 7), bagging (M 9), RF (M 10), AdaBoost (M 11), and SGB (M 12)] have produced a high acceptable performance where its accuracy is ranging from 9.714% to 11.008 %. The ensemble learning methods can effectively deal with the problems of high-dimension data, complex data structures, and small sample size. Bagging algorithms can increase generalization by decreasing variance error (Breiman 1998) while boosting can improve generalization by decreasing bias error (Schapire et al. 1998). Ensemble methods can effectively handle continues, categorical, and dummy features with missing values. However, ensemble methods may increases model complexity which decreases the model interpretability (Kuncheva 2004). RF (M 10) is a robust algorithm against noisy data or big data than the DT (M 16) algorithm (Breiman, 1996; Dietterich, 2000). However, RF algorithm is unable to interpret the importance of features or the mechanism of producing the results.

DNNs (M 15) produces 12.059% MAPE less than all the developed MLP (M 4, M 5, and M 8). Accordingly, DNNs provide bad performances with a small dataset. Conversely, deep learning and DNNs can produce the most accurate performance with high dimension data (LeCun et al. 2015). An alternative to the black box nature of ANNs and DNNs, DT generates logic statements and interpretable rules which can be used for identifying the importance of data features (Perner et al. 2001). Another advantage of DT is avoiding the curse dimensionality and providing a high-performance computing efficiency through its splitting procedure (Prasad et al. 2006). However, DT is producing unsatisfactory performance in time series, noisy, or nonlinear data (Curram and Mingers 1994). Although DT (CART) is inherently used as a based learner for the ensemble methods, DT (M 16) produces 12.488 % MAPE less than all developed ensemble methods (M1, M7, M9, M10, M11, and M12). Therefore, Ensemble methods produce better performance than a single learning algorithm. Moreover, ensemble methods can effectively handle missing values and noisy data due to scalability.

Ensemble methods and data transformation play an important role in prediction accuracy. However, the main gap of the previous models is lacking the uncertainty modeling to the prediction cost model. Therefore, Fuzzy logic theory has been conducted to maintain uncertainty concept through fuzzy logic model (M 17) and hybrid fuzzy model (M 20). The number of generated rules by the fuzzy genetic model (M 17) are 63 rules and the MAPE is 14.7%. On the other hand, a traditional fuzzy logic model (M 20) has been built based on the experts 'experience where a total of 190 rules are generated to cover all the possible combinations of the fuzzy system and MAPE is 26.3 %. Moreover, the fuzzy rules (IF-Then rules) generated by experts have redundant rules which can be deleted to improve the model computation and performance. Moreover, the expert's knowledge cannot cover all combination to represent all possible rules (2401 rules). In addition, the generation of the experts' rules is time and effort consuming process. Consequently, hybrid fuzzy systems are more effective than traditional fuzzy logic system. Although the prediction accuracy of the fuzzy genetic model (M 17) and the fuzzy logic model (M 20) is 14.7 % and 26.3 %, respectively, the fuzzy model would produce more reliable prediction results due to taking uncertainty into account. However, the traditional fuzzy model gives unacceptable accuracy of 26.3% MAPE (Peurifoy and Oberlender 2002). Therefore, maintaining uncertainty decrease the predictive model accuracy.

CBR (M 18) produces acceptable low accuracy of 17.3% MAPE. The advantage of CBR is dealing with a vast amount of data where all past cases and new cases are stored in database techniques (Kim and Kang 2004). Moreover, Finding similarities and similar cases improve the reliability and confidence in the output. Hybrid models can be incorporated to CBR to enhance the performance of CBR such as applying GA and decision tree to optimize attributes weights and applying regression analysis for the revising process. SVM can be applied for both regression and classification tasks. SVM (M 19) produces unacceptable accuracy of 21.217% MAPE (Peurifoy and Oberlender 2002). Finally, Table.2 summarizes strengths and weakness of each developed model.

**Table.1: Accuracy of the developed algorithms.**

| Notation | Algorithm / model | Algorithm type: supervised regression | MAPE % | MAPE % categorization | $R^2$ | $R^{*2}$ |
|---|---|---|---|---|---|---|
| M 1 | **XGBoost** | **Ensemble methods** | **9.091** | **below 10** | **0.931** | **0.929** |
| M 2 | **Quadratic regression*** | **MRA** | **9.120** | **below 10** | **0.857** | **0.851** |
| M 3 | **Plain regression*** | **MRA** | **9.130** | **below 10** | **0.803** | **0.796** |
| M 4 | **Quadratic MLP*** | **ANNs** | **9.200** | **below 10** | **0.904** | **0.902** |
| M 5 | **Plain MLP*** | **ANNs** | **9.270** | **below 10** | **0.913** | **0.912** |
| M 6 | **Semilog regression*** | **MRA** | **9.300** | **below 10** | **0.915** | **0.910** |
| M 7 | **Extra Trees** | **Ensemble methods** | **9.714** | **below 10** | **0.948** | **0.947** |
| M 8 | **Natural log MLP*** | **ANNs** | **10.230** | **below 20** | **0.905** | **0.910** |
| M 9 | **Bagging** | **Ensemble methods** | **10.246** | **below 20** | **0.914** | **0.911** |
| M 10 | **RF** | **Ensemble methods** | **10.503** | **below 20** | **0.916** | **0.913** |
| M 11 | **AdaBoost** | **Ensemble methods** | **10.679** | **below 20** | **0.875** | **0.871** |
| M 12 | **SGB** | **Ensemble methods** | **11.008** | **below 20** | **0.926** | **0.924** |
| M 13 | **Reciprocal regression*** | **MRA** | **11.200** | **below 20** | **0.814** | **0.801** |
| M 14 | **Power (2) regression*** | **MRA** | **11.790** | **below 20** | **0.937** | **0.931** |
| M 15 | **DNNs** | **ANNs** | **12.059** | **below 20** | **0.785** | **0.779** |
| M 16 | **DT** | **Tree model** | **12.488** | **below 20** | **0.886** | **0.883** |
| M 17 | **Genetic-Fuzzy** | **Hybrid model** | **14.700** | **below 20** | **0.863** | **0.857** |
| M 18 | **CBR** | **Case based** | **17.300** | **below 20** | **0.859** | **0.852** |
| M 19 | **SVM** | **Kernel based** | **21.217** | **unacceptable** | **0.136** | **0.133** |
| M 20 | **Fuzzy** | **Fuzzy theory** | **26.300** | **unacceptable** | **0.857** | **0.851** |

* : (Elmousalami et at. 2018 b)

**Table.2: characteristics of the developed algorithms.**

| | Strengths | weaknesses | interpretation | uncertainty | noisy data |
|---|---|---|---|---|---|
| **M 1** | high scalability, handing missing values, high accuracy, low computational cost | no uncertainty and interpretation | no | no | yes |
| **M 2** | more accurate then plain regression, hand nonlinearity of data | prone to overfitting | yes | no | no |
| **M 3** | Works on small size of dataset | linear assumptions | yes | no | no |
| | | | | | |
| **M 4** | High accuracy, handling complex patterns | Black box nature, need sufficient data for training | no | no | no |
| **M 5** | High accuracy, handling complex patterns | Black box nature | no | no | no |
| **M 6** | producing better results than plain regression | Unable to capture complex patterns | yes | no | no |
| **M 7** | handing data randomness | Black box nature, sufficient data | no | no | yes |
| **M 8** | producing better results than plain MLP | Black box nature, sufficient data | no | no | no |
| **M 9** | providing higher performance than a single algorithm | depending on other algorithms performance | no | no | yes |
| **M 10** | Accurate and high performance on many problems including non linear | No interpretability, overfitting can easily occur, need to choose the number of trees | no | no | yes |
| **M 11** | high scalability, and high adaptability | depends on other algorithms performance | no | no | yes |
| **M 12** | handing difficult examples | high sensitive to noisy data | no | no | yes |
| **M 13** | handing data nonlinearity and training small sample size | Unable to capture complex patterns | yes | no | no |
| **M 14** | handing data nonlinearity and training small sample size | Unable capture complex patterns | yes | no | no |
| **M 15** | Capturing complex patterns, processing big data and high performance computing (HPC) | Sufficient training data and high cost computation | no | no | no |
| **M 16** | working on both linear / nonlinear problems , and producing logical expressions | Poor results on too small datasets, overfitting can easily occur | yes | no | no |
| **M 17** | Handing uncertainty and more accurate then fuzzy model | more complex that fuzzy model and needs more computational resources | yes | yes | no |
| **M 18** | Handling small data sets, simple and needs less computational time | poor performance and accuracy where the optimal case cannot be retrieved | yes | no | no |
| **M 19** | Easily adaptable, works very well on nonlinear problems, not biased by outliers | Compulsory to apply feature scaling, not well known, more difficult to understand | no | no | no |
| **M 20** | handing uncertainty | low accuracy | yes | yes | no |

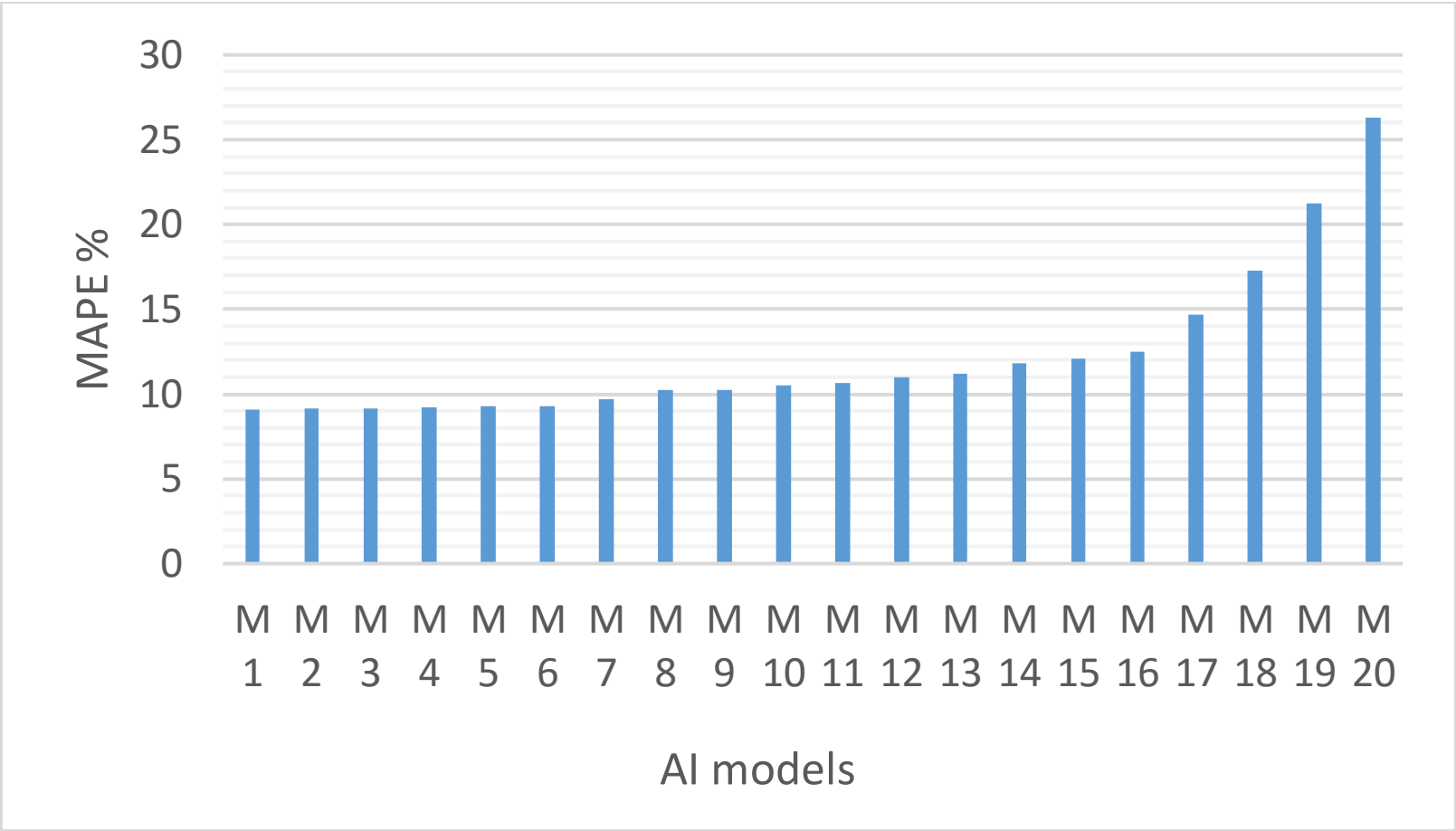

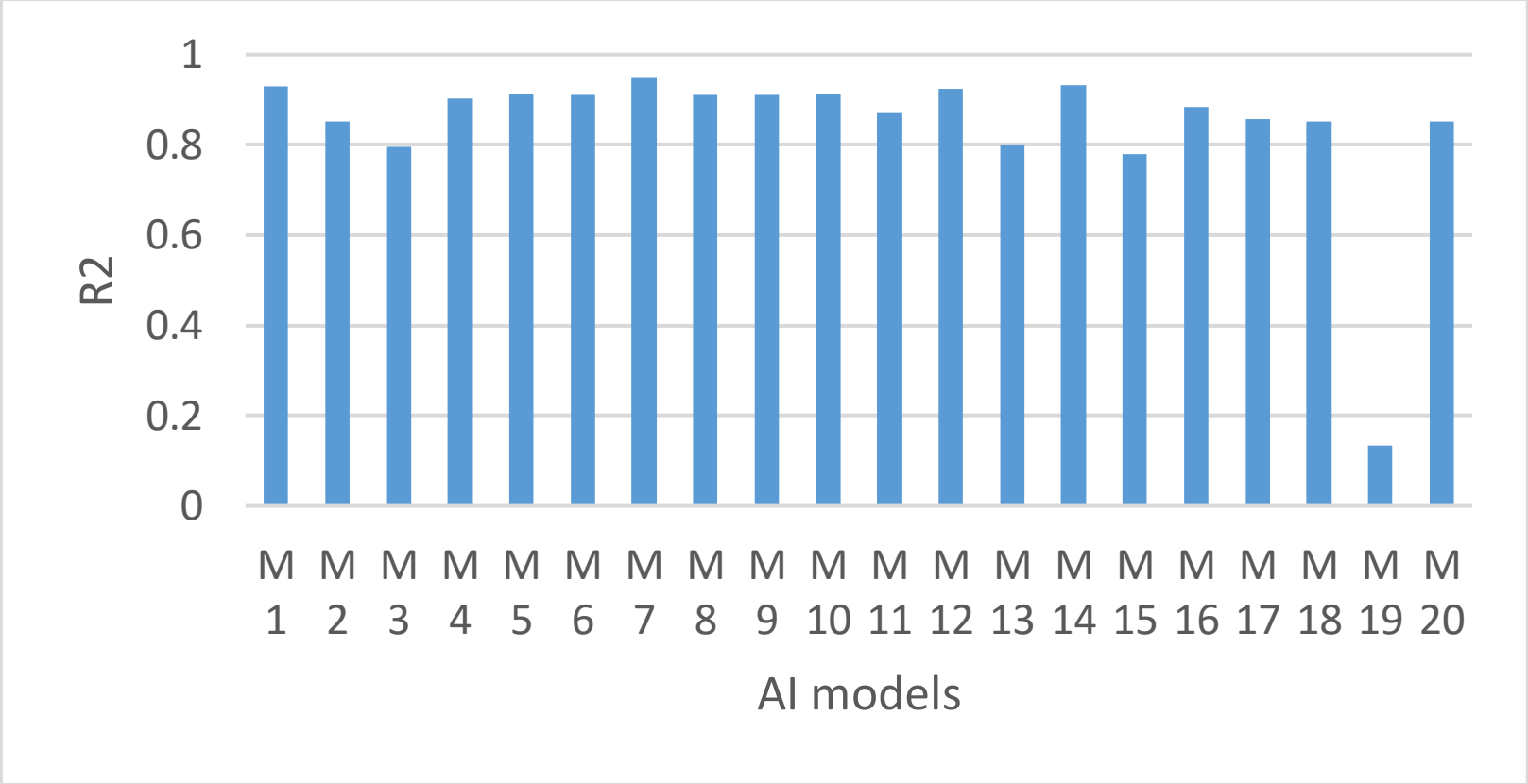

Fig.7: MAPE and $R^{*2}$ for all algorithms.

**Conclusion**

This paper presents a comparison of AI techniques to develop a reliable conceptual cost prediction model. Twenty one machine learning models are developed utilizing tree based models, ensemble methods, fuzzy systems, CBR, ANNs SVM, and transformed regression models. The accuracy of the developed models are tested from two perspectives: MAPE and adjusted $R^2$. The results show that the most accurate and suitable method is XGBoost with 9.091% and 0.929 for MAPE and adjusted $R^2$, respectively. The study emphasis the importance of ensemble methods for improving the prediction accuracy, handling noisy and missing data. However, the key limitation of the ensemble methods in inability to interpret the producing results. In addition, Decision tree algorithm and ensemble methods can provide an alternative technique to many ML algorithms such as multiple regression analysis and ANNs.

The conceptual cost estimate is conducted under uncertainty. Therefore, this study recommended using fuzzy theory such as FL and to develop a hybrid model based on FL to obtain uncertainty nature for the developed model and produce more reliable performance. In addition, the study highlights the main problem for fuzzy modeling which is fuzzy rules generation. This study has discussed the importance of the hybrid fuzzy model methodologies to generate rules such as fuzzy genetic model. Therefore, this study recommends developing an automated hybrid fuzzy rules models than traditional fuzzy models. The Fusion of the AI techniques is called hybrid intelligent systems where Zadeh (1994) has predicted that the hybrid intelligent systems will be the way of the future.

It is recommended to develop more than one cost prediction model such as the regression model, ANNs, FL, ensemble methods or CBR model. As a result, the researcher can compare the results of the developed models and set a benchmark to select the most accurate model. In addition, the comparisons of the developed models enhance the quality of cost estimate and the decision based on it (Amason 1996). Moreover, this paper will provide the comprehensive knowledge needed to develop a reliable parametric cost model at the conceptual stage of the project. However, Automation the cost models are prone to many machine learning problems such as overfitting issues, and hyper-parameter selection.